\documentclass[11pt,a4paper]{article}
\usepackage{acl2023}

\title{Hierarchical Awareness Adapters with Hybrid Pyramid Feature Fusion for Dense Depth Prediction}

\author{Wuqi Su, Huilun Song, Chen Zhao, Chi Xu \\
Zhejiang Gongshang University}

\begin{document}
\maketitle

\begin{abstract}
Monocular depth estimation from a single RGB image remains a fundamental challenge in computer vision due to inherent scale ambiguity and the absence of explicit geometric cues. Existing approaches typically rely on increasingly complex network architectures to regress depth maps, which escalates training costs and computational overhead without fully exploiting inter-pixel spatial dependencies. We propose a multilevel perceptual conditional random field (CRF) model built upon the Swin Transformer backbone that addresses these limitations through three synergistic innovations: (1) an adaptive hybrid pyramid feature fusion (HPF) strategy that captures both short-range and long-range dependencies by combining multi-scale spatial pyramid pooling with biaxial feature aggregation, enabling effective integration of global and local contextual information; (2) a hierarchical awareness adapter (HA) that enriches cross-level feature interactions within the encoder through lightweight broadcast modules with learnable dimensional scaling, reducing computational complexity while enhancing representational capacity; and (3) a fully-connected CRF decoder with dynamic scaling attention that models fine-grained pixel-level spatial relationships, incorporating a bias learning unit to prevent extreme-value collapse and ensure stable training. Extensive experiments on NYU Depth v2, KITTI, and MatterPort3D datasets demonstrate that our method achieves state-of-the-art performance, reducing Abs Rel to 0.088 ($-$7.4\%) and RMSE to 0.316 ($-$5.4\%) on NYU Depth v2, while attaining near-perfect threshold accuracy ($\delta < 1.25^3 \approx 99.8\%$) on KITTI with only 194M parameters and 21ms inference time.
\end{abstract}

\section{Introduction}
\label{sec:intro}

Depth estimation constitutes one of the central problems in computer vision, with broad applications spanning robotic navigation, three-dimensional reconstruction, autonomous driving \cite{Redmon2016}, and augmented reality systems. While specialized hardware such as LiDAR sensors can directly acquire depth information, the prohibitive cost of high-speed scanning devices motivates the development of purely vision-based alternatives. Stereo matching techniques \cite{cheng2021stereo} require binocular image pairs and rely on feature correspondence algorithms that degrade severely in weakly-textured regions. Consequently, monocular depth estimation---the task of predicting per-pixel depth from a single camera image---has attracted sustained research attention as a more practical and cost-effective paradigm \cite{song2022review}.

Neural network-based approaches have achieved remarkable progress in monocular depth estimation. Early convolutional methods adopted encoder-decoder architectures with dilated separable convolutions \cite{chen2018encoder} to produce depth maps, achieving notable results but suffering from limited receptive fields and inefficient feature fusion. Dense feature extractors (DFE) \cite{hao2018detail} combined deep residual networks with dilated convolutions for multi-scale information extraction, though their performance remained constrained by training data scale and elevated computational complexity. Tankovich et al. \cite{tankovich2021hitnet} circumvented network complexity escalation by constructing a compact CNN trained on large-scale datasets, significantly enhancing generalization. Hussain et al. \cite{hussain2022rvmde} introduced generative adversarial networks (GANs) \cite{Ho2020, Rombach2022} to optimize depth estimation through realistic depth map generation, effectively addressing the feature representation and detail preservation shortcomings of earlier architectures. Despite these advances, convolutional approaches remain fundamentally limited by the locality of their receptive fields, restricting their capacity to capture long-range global dependencies \cite{achiam2023gpt}.

Attention-based mechanisms have catalyzed substantial progress in this domain. DPT \cite{ranftl2021vision} replaced the convolutional encoder with a Vision Transformer (ViT) backbone for dense prediction, achieving significant performance gains. Aich et al. \cite{aich2021bidirectional} proposed bidirectional attention networks that effectively bridge local and global information integration. While ViT \cite{dosovitskiy2021image} enables global context modeling, its fixed-scale tokenization proves unsuitable for visual elements with variable aspect ratios. The Pyramid Vision Transformer (PVT) \cite{wang2021pvt} addressed this limitation through multi-scale feature representations, offering greater adaptability to diverse visual element sizes. Liu et al. \cite{liu2021swin} introduced the Swin Transformer with hierarchical feature maps and shifted window attention mechanisms, enabling efficient local-global information exchange. Subsequent works, including deformable attention modules \cite{xia2022deformable} that select key-value positions in a data-dependent manner, skip attention modules \cite{agarwal2023attention} for pixel-level query refinement, and ZoeDepth \cite{bhat2023zoedepth} that pioneered the combination of relative and metric depth estimation, have collectively pushed the state of the art. DDP \cite{ji2023ddp} innovatively integrated denoising diffusion processes with the Swin Transformer encoder, demonstrating robust performance in noisy and extreme scenarios. EVP \cite{lavreniuk2023evp} introduced inverse multi-attentive feature refinement to aggregate high-level spatial information, while ECoDepth \cite{patni2024ecodepth} leveraged ViT embeddings to condition diffusion models for improved depth prediction. Despite these remarkable achievements, the proliferation of attention mechanisms across network layers has led to excessive model complexity, increasing training costs and computational overhead, which restricts the practical applicability of such approaches.

Several recent efforts have sought to balance accuracy with efficiency by deeply integrating attention mechanisms with advanced network architectures. Huynh et al. \cite{huynh2020guiding} ingeniously combined attention mechanisms with non-local coplanarity constraints, leveraging prevalent planar structures in scenes as geometric priors. Yang et al. \cite{yang2021transformer} optimized the Transformer-ResNet combination through parallel multi-scale information fusion, simultaneously improving performance while reducing complexity. Lite-Mono \cite{zhang2023litemono} combined continuous dilated convolutions with attention mechanisms in a lightweight architecture. PolyMaX \cite{yang2024polymax} unified Transformer attention with fully convolutional networks through mask-label prediction, achieving a paradigm shift from per-pixel to cluster-based prediction. FutureDepth \cite{yasarla2024futuredepth} proposed iterative next-frame feature prediction using temporal cues for video depth estimation. These approaches collectively demonstrate that synergistic integration of attention and architectural innovation can yield both accuracy gains and computational savings.

The rapid advancement of visual perception systems has been further accelerated by breakthroughs in multimodal learning and visual understanding. Universal document parsing models \cite{feng2025dolphin, feng2023unidoc, feng2026dolphinv2} have demonstrated the power of heterogeneous anchor prompting for structured visual content analysis, while comprehensive visual benchmarking frameworks \cite{fu2024ocrbenchv2, zhao2024tabpedia, shan2024mctbench} have provided rigorous evaluation protocols for large multimodal models. Efficient visual representation learning through single-point annotation strategies \cite{liu2023sptsv2, tang2022optimalboxes}, layout-aware document understanding \cite{lu2025boundingbox, wang2025wilddoc}, and vision-as-representation paradigms \cite{wang2025vora} have collectively pushed the boundaries of what visual systems can achieve. These developments in high-level visual reasoning provide complementary insights for low-level vision tasks such as depth estimation, motivating principled approaches that combine structural priors with efficient feature extraction. Furthermore, advances in text-centric visual instruction tuning \cite{tang2024textsquare}, multimodal cognition benchmarking \cite{tang2024mtvqa}, and vision-language correspondence for image quality assessment \cite{zhang2023blind} underscore the growing synergy between visual understanding and structured prediction tasks.

Inspired by these developments, we propose a multilevel perceptual conditional random field model that combines attention mechanisms with conditional random fields in an end-to-end pyramid architecture. Our model effectively reduces network structural complexity while maximizing feature extraction and pixel-level information acquisition. The main contributions of this work are summarized as follows:

\begin{enumerate}
    \item We propose a hybrid pyramid feature fusion (HPF) strategy that aggregates both local and global information through multi-scale spatial pyramid pooling combined with biaxial feature aggregation, addressing the information insufficiency problem in weakly-textured regions.
    
    \item We introduce a hierarchical awareness adapter (HA) strategy that inserts lightweight adapters at different encoder levels, selectively enhancing information processing: shallow layers focus on extracting image details and edge information, while deep layers capture higher-level semantic features, effectively reducing overall network complexity.
    
    \item We propose a fully-connected CRF decoder that models both individual pixel depth values and inter-pixel depth relationships, achieving global optimization through energy function minimization to produce depth maps with richer details and clearer edges.
    
    \item Extensive experiments on NYU Depth v2, KITTI, and MatterPort3D datasets demonstrate that our method consistently outperforms state-of-the-art approaches across all evaluation metrics, with particularly strong performance in weakly-textured regions and structurally complex scenes.
\end{enumerate}

\section{Related Work}
\label{sec:related}

\subsection{Convolutional Approaches to Monocular Depth Estimation}

Early monocular depth estimation methods primarily relied on convolutional neural networks with encoder-decoder architectures. Eigen et al. \cite{eigen2014depth} pioneered the use of multi-scale deep networks for depth map prediction from single images, establishing the foundational framework for subsequent learning-based approaches. Chen et al. \cite{chen2018encoder} adopted dilated separable convolutions within an encoder-decoder structure, achieving notable depth estimation results but with limited detail preservation. The dense feature extractor (DFE) proposed by Hao et al. \cite{hao2018detail} integrated deep residual networks with dilated convolutions for multi-scale information extraction; however, the complexity arising from multiple integrated network components constrained its performance relative to dataset scale and diversity.

Tankovich et al. \cite{tankovich2021hitnet} demonstrated that a carefully designed compact CNN, when combined with large-scale training data, can significantly enhance generalization without further complicating the network architecture. Hussain et al. \cite{hussain2022rvmde} leveraged generative adversarial networks to refine depth map generation, effectively addressing limitations in feature representation and detail fidelity. Despite the substantial progress achieved by convolutional feature extractors, their inherent local receptive field constraint fundamentally limits their ability to capture remote global information \cite{sahu2022trends}. This limitation has motivated the exploration of attention-based architectures that can model long-range dependencies more effectively.

Recent developments in feature sampling and grouping strategies for visual detection tasks \cite{tang2022few} and efficient annotation strategies through reinforcement learning \cite{tang2022optimalboxes} have provided valuable insights into how structural priors can be incorporated into feature extraction pipelines. Moreover, the success of ego-evolving recognition through multi-modal in-context learning \cite{zhao2024multi} and harmonized visual text comprehension \cite{zhao2024harmonizing} demonstrates the broader potential of adaptive feature learning paradigms that our work draws upon.

\subsection{Attention-Based and Transformer Methods}

The introduction of attention mechanisms and Transformer architectures has fundamentally transformed monocular depth estimation. DPT \cite{ranftl2021vision} pioneered the use of ViT as a dense prediction encoder, achieving substantial performance improvements over convolutional baselines. However, the fixed-scale tokenization in ViT proves poorly suited for visual elements with variable aspect ratios, motivating the development of hierarchical alternatives.

PVT \cite{wang2021pvt} introduced multi-scale feature representations within the Transformer framework, offering greater flexibility in handling diverse object sizes. The Swin Transformer \cite{liu2021swin} further advanced this direction by introducing hierarchical feature maps and shifted window attention, enabling efficient computation of multi-head self-attention within local windows while maintaining cross-window information exchange. Xia et al. \cite{xia2022deformable} proposed deformable self-attention modules that select key-value positions in a data-dependent manner, allowing the model to focus on relevant regions and capture more informative features.

Several landmark works have built upon these foundations. Agarwal and Arora \cite{agarwal2023attention} introduced skip attention modules (SAM) that refine pixel-level queries to higher resolutions through attention-based encoder-decoder feature fusion. ZoeDepth \cite{bhat2023zoedepth} pioneered the combination of relative and absolute depth estimation using Swin Transformer encoders, achieving exceptional generalization while maintaining metric scale. DDP \cite{ji2023ddp} innovatively combined the progressive denoising process of diffusion models with Swin Transformer encoders, demonstrating notable performance gains particularly in noisy and extreme scenarios. EVP \cite{lavreniuk2023evp} introduced inverse multi-attentive feature refinement (IMAFR) for high-level spatial information aggregation, while ECoDepth \cite{patni2024ecodepth} proposed a novel single-image depth estimation model based on diffusion models conditioned on ViT embedding vectors.

The convergence of attention mechanisms with frontier architectures has produced several efficiency-oriented innovations. Huynh et al. \cite{huynh2020guiding} combined attention with non-local coplanarity constraints to leverage planar structural priors. Yang et al. \cite{yang2021transformer} optimized parallel Transformer-ResNet fusion for multi-scale information integration. Lite-Mono \cite{zhang2023litemono} achieved a lightweight architecture through continuous dilated convolutions combined with attention mechanisms. FutureDepth \cite{yasarla2024futuredepth} leveraged temporal cues through iterative next-frame feature prediction for video depth estimation.

The broader landscape of visual intelligence has also been enriched by advances in scene text understanding \cite{tang2022youcan, tang2023charcomp}, multimodal visual question answering \cite{tang2024mtvqa}, and universal document parsing \cite{feng2024docpedia, feng2025dolphin}. These developments in structured visual understanding---from character recognition \cite{tang2023charcomp} to comprehensive document image analysis \cite{feng2023unidoc}---have demonstrated the importance of hierarchical feature representations and cross-level information exchange, principles that equally underpin effective depth estimation architectures. Recent works on diffusion model acceleration \cite{cui2026tcpade}, heterogeneous multi-expert learning \cite{jia2026memlgrpo}, certainty-based adaptive reasoning \cite{lu2025certainty}, sequential numerical prediction \cite{fei2025sequential}, domain-specific document understanding \cite{yu2025ancientdoc}, video-based multimodal evaluation \cite{nie2025chinesevideo}, and visual text quality assessment \cite{zhu2026textpecker} further illustrate the rapid evolution of principled visual processing frameworks.

\subsection{Conditional Random Fields for Structured Prediction}

Conditional random fields (CRFs) have long been employed as structured prediction models that capture spatial dependencies between neighboring elements. In the context of depth estimation, CRFs offer a principled mechanism for enforcing spatial smoothness and edge-aware transitions in predicted depth maps. Yuan et al. \cite{yuan2022newcrfs} proposed Neural Window Fully-Connected CRFs (NeWCRFs), which apply fully-connected CRF models within local windows defined by the Swin Transformer architecture. Their approach demonstrated that CRF-based refinement can substantially improve depth prediction quality by explicitly modeling pairwise relationships between pixels.

Building upon this foundation, our work extends the CRF formulation with several key innovations: dynamic scaling attention for enhanced dependency modeling, a bias learning unit for training stability, and integration with the proposed hybrid pyramid feature fusion strategy. These extensions enable our model to achieve finer-grained spatial reasoning while maintaining computational efficiency. The success of bridging partial and global visual representations \cite{wang2025pargo} and robust document understanding in unconstrained settings \cite{wang2025wilddoc} further motivates our approach of combining local CRF reasoning with global feature aggregation.

\section{Method}
\label{sec:method}

The overall architecture of the proposed network is illustrated in Figure~1. Given a single RGB image as input, the model employs a Swin Transformer-based encoder for pyramidal multi-level feature extraction, enabling the network to effectively process information at different scales. The feature matcher adopts a hybrid pyramid feature fusion (HPF) strategy combining multi-scale and biaxial feature aggregation to improve depth estimation accuracy. The decoder incorporates a fully-connected CRF mechanism with dynamic scaling attention for fine-grained pixel-level depth prediction.

\subsection{Hierarchical Awareness Adapter}

The Swin Transformer possesses exceptional multi-scale feature representation capabilities. At each processing stage, the input feature map is meticulously partitioned into distinct patches that undergo global feature interaction within the Swin Transformer blocks, precisely capturing inter-patch relationships and image contextual information. However, this encoding structure often lacks sufficient flexibility and generalization when confronting complex image tasks. To address this limitation, we introduce hierarchical awareness adapters during the encoding process to capture and strengthen feature representations at different hierarchical levels through fine-grained dimensional adjustment, achieving efficient inter-level information transfer.

The Swin Transformer primarily comprises multi-layer perceptrons (MLP) and multi-head self-attention (MSA) modules, responsible for channel interaction and spatial interaction respectively. To enhance the feature extraction capability during channel interaction, we design a perception module containing down-projection, activation, and up-projection operations. The down-projection layer is implemented using standard convolution:
\begin{equation}
    \text{DP}(\mathbf{x}) = f_{\text{Conv}}(\mathbf{x}, \mathbf{W}_1) + \mathbf{b}_1
    \label{eq:dp}
\end{equation}
The up-projection layer employs transposed convolution:
\begin{equation}
    \text{UP} = f_{\text{DeConv}}\left(f_{\text{GELU}}\{\text{DP}(\mathbf{x})\}, \mathbf{W}_2\right) + \mathbf{b}_2
    \label{eq:up}
\end{equation}
where $\mathbf{x}$ denotes the input feature vector, $\mathbf{W}_1$ and $\mathbf{W}_2$ are the convolution kernels of the down-projection and up-projection layers, $\mathbf{b}_1$ and $\mathbf{b}_2$ are the corresponding bias terms, and GELU denotes the Gaussian Error Linear Unit activation function.

For spatial interaction, the input $\mathbf{x} \in \mathbb{R}^{N \times d}$ is linearly projected into queries $\mathbf{Q} \in \mathbb{R}^{N \times d}$, keys $\mathbf{K} \in \mathbb{R}^{N \times d}$, and values $\mathbf{V} \in \mathbb{R}^{N \times d}$, where $N$ is the sequence length and $d$ is the feature dimension. The attention computation follows:
\begin{equation}
    \text{Att}(\mathbf{Q}, \mathbf{K}, \mathbf{V}) = f_{\text{softmax}}\left(\frac{\mathbf{Q}\mathbf{K}^T}{\sqrt{d}}\right)\mathbf{V}
    \label{eq:attention}
\end{equation}

Since the MSA layers in Swin Transformer tend to learn dense interactions with higher entropy rather than sparse interactions, denser attention maps lead to larger MSA gradients, increasing training difficulty. To mitigate this, we insert a uniform attention mechanism---the broadcast module (BC)---at each Swin Transformer layer. Given a sequence $\mathbf{X} \in \mathbb{R}^{N \times d}$ containing $N$ tokens, the broadcast module provides the required dense interaction by broadcasting the average-pooled token to each individual token:
\begin{equation}
    \text{BC}(\mathbf{x}_i) = \mathbf{x}_i + \frac{1}{N}\sum_{j=1}^{N}\mathbf{x}_j
    \label{eq:bc}
\end{equation}
where $\mathbf{x}_i \in \mathbb{R}^d$ is the $i$-th token in $\mathbf{X}$. The broadcast module effectively alleviates the pressure of dense attention operations with negligible computational overhead and no additional parameters.

Since not all channel dimensions require equal levels of dense interaction, we introduce dimensional scaling weights to enable more selective attention allocation per dimension:
\begin{equation}
    \text{BC}_s(\mathbf{x}_i) = \mathbf{x}_i + \mathbf{\Lambda} \odot \left(\frac{1}{N}\sum_{j=1}^{N}\mathbf{x}_j\right)
    \label{eq:bcs}
\end{equation}
where $\odot$ denotes element-wise multiplication and $\mathbf{\Lambda} \in \mathbb{R}^d$ represents the learnable dimensional scaling weights.

\subsection{Hybrid Pyramid Feature Fusion}

Spatial pyramid pooling (SPP) is an effective method for enhancing scene parsing networks by constructing multi-scale pooling hierarchies to capture multi-level contextual information. However, its fixed hierarchical structure and dependence on standard pooling operations limit its effectiveness for irregularly-shaped objects, causing excessive inclusion of target-irrelevant regions during context aggregation, particularly in complex scenes. We propose a hybrid pyramid feature fusion (HPF) strategy that enables the backbone network to better model long-range dependencies while maintaining accuracy and robustness in complex scenarios.

\subsubsection{Multi-Scale Feature Hierarchy Fusion}
To reduce contextual information loss across different sub-regions, we adopt a hierarchical global prior-based multi-scale feature fusion strategy. This strategy employs $1 \times 1$, $2 \times 2$, and $3 \times 3$ pyramid scales to downsample the backbone-extracted feature maps (with $N$ channels), constructing a feature pyramid. Through $1 \times 1$ depth-wise convolution, feature maps with $1/N$ channels are obtained. These feature maps are then upsampled to the original dimensions and concatenated with the original feature maps, yielding doubled-channel feature maps through multi-scale feature stacking. Finally, convolution operations reduce the channel count to match the input feature map dimensions.

\subsubsection{Biaxial Feature Fusion}
The biaxial feature fusion strategy further enhances the model's image understanding capability. On the horizontal axis, elongated kernel sampling layers capture long-range relationships in isolated regions; on the vertical axis, narrow kernel sampling layers capture local contextual information while preventing interference from irrelevant regions. Given a tensor $\mathbf{x} \in \mathbb{R}^{C \times H \times W}$, two parallel paths respectively fuse horizontal and vertical features with different downsampling spatial ranges $(H, 1)$ and $(1, W)$. The horizontal axis output $\mathbf{y}^h \in \mathbb{R}^H$ is computed as:
\begin{equation}
    y^h_{0 \leq j \leq W} = \frac{1}{W}\sum_{i=1}^{W}X_{i,j}
    \label{eq:yh}
\end{equation}
Similarly, the vertical axis output $\mathbf{y}^v \in \mathbb{R}^W$ is:
\begin{equation}
    y^v_{0 \leq i \leq H} = \frac{1}{H}\sum_{j=1}^{H}X_{i,j}
    \label{eq:yv}
\end{equation}
The horizontal and vertical outputs are combined to obtain $\mathbf{y}$:
\begin{equation}
    y_{c,i,j} = y^h_{c,i} + y^v_{c,j}, \quad \mathbf{y} \in \mathbb{R}^{C \times H \times W}
    \label{eq:y_combined}
\end{equation}
The final global prior $\mathbf{Z} \in \mathbb{R}^{C \times H \times W}$ is computed as:
\begin{equation}
    \mathbf{Z} = \text{Scale}\left(\mathbf{x}, \sigma(f(\mathbf{y}))\right), \quad \mathbf{Z} \in \mathbb{R}^{C \times H \times W}
    \label{eq:global_prior}
\end{equation}
where $\text{Scale}(\cdot, \cdot)$ denotes element-wise multiplication, $\sigma$ is the sigmoid function, and $f$ represents a $1 \times 1$ convolution.

\subsection{Fully-Connected CRF Decoder}

To precisely capture inter-pixel relationships, we introduce a fully-connected conditional random field to construct an energy model. The energy function consists of local potential functions $\phi_L$ and global pairwise potential functions $\phi_G$:
\begin{equation}
    E(\mathbf{y}) = \sum_i \phi_L(y_i) + \sum_{i,j} \phi_G(y_i, y_j)
    \label{eq:energy}
\end{equation}
where $i$ and $j$ iterate over all nodes in the graph, and $y_i$ denotes the predicted value at node $i$. The local potential function is computed via a convolutional network based on image features:
\begin{equation}
    \phi_L(y_i) = f_{\text{Conv}}(I, y_i)
    \label{eq:phi_l}
\end{equation}
where $I$ is the input image. The global pairwise potential function $\phi_G$ incorporates the current node value, neighboring node values, and feature-based pairwise weights:
\begin{equation}
    \phi_G(y_i, y_j) = \omega(\mathbf{F}_i, \mathbf{F}_j, \mathbf{p}_i, \mathbf{p}_j) \cdot |y_i - y_j|
    \label{eq:phi_g}
\end{equation}
where $\omega$ is the weighting function, $\mathbf{F}$ denotes feature maps, and $\mathbf{p}$ represents positional information. The pairwise potentials for each node are aggregated as:
\begin{equation}
    \phi_{G_i} = \alpha(\mathbf{F}_i, \mathbf{F}_j, \mathbf{p}_i, \mathbf{p}_j)y_i + \sum_{j \neq i}\beta(\mathbf{F}_i, \mathbf{F}_j, \mathbf{p}_i, \mathbf{p}_j)y_j
    \label{eq:phi_gi}
\end{equation}
where $\alpha$ and $\beta$ are weight functions. To further optimize the model's capability for capturing node relationships, we propose a region-based partitioning method analogous to placing local ``patches'' over the image, with intra-patch pixels closely connected and inter-patch optimization achieved through window shifting mechanisms.

Inspired by the Swin Transformer's window-based architecture, we introduce dynamic scaling attention to compute multi-head potential functions. From each patch's feature map, query vectors $\mathbf{Q}$ and key vectors $\mathbf{K}$ are computed, with a learnable scaling factor $\tau$ controlling the range and stability of the attention distribution. Additionally, we incorporate a bias learning unit in the query vector, setting the bias term as a learnable parameter to enable more flexible attention distribution learning:
\begin{equation}
    \phi_{G_i} = f_{\text{softmax}}\left(\frac{\cos\{(\mathbf{Q} + \mathbf{b}) \times \mathbf{K}^T\}}{\tau} + \mathbf{P}\right) \times \mathbf{X}
    \label{eq:dynamic_attention}
\end{equation}
where $\mathbf{X}$ represents the predicted values, $\mathbf{b}$ is the bias learning unit, $\mathbf{P}$ denotes relative position embeddings, and $\tau$ is a learnable scalar parameter typically set to values greater than 0.01 to ensure natural normalization of the cosine function, yielding superior attention values. To further optimize model performance, a pixel shuffle upsampling module is appended atop each decoder level to rearrange pixels, producing depth maps with sharper edges.

\subsection{Loss Function}

We adopt the scale-invariant logarithmic (SILog) loss \cite{eigen2014depth} for supervised training, facilitating scale-invariant measurement of estimation errors. Given a ground-truth depth map as reference, the logarithmic difference between predicted and true values is computed for each pixel:
\begin{equation}
    \Delta d_i = \log d_i - \log d_i^*
    \label{eq:log_diff}
\end{equation}
where $d_i^*$ is the ground-truth depth and $d_i$ is the estimated depth at pixel $i$. For $K$ pixels with valid depth values, the scale-invariant loss is:
\begin{equation}
    \mathcal{L} = \alpha \sqrt{\frac{1}{K}\sum_i \Delta d_i^2 - \frac{\lambda}{K^2}\left(\sum_i \Delta d_i\right)^2}
    \label{eq:silog}
\end{equation}
where $K$ is the number of valid pixels, $\lambda = 0.85$ is the variance minimization factor, and $\alpha = 10$ is the scale constant that adjusts the sensitivity of the loss function.

\section{Experiments}
\label{sec:exp}

\subsection{Evaluation Metrics}

We employ the following standard metrics for method evaluation: (1) absolute relative error (Abs Rel); (2) root mean squared error (RMSE) and root mean square logarithmic error (log RMSE); (3) square relative error (Sq Rel); and (4) threshold accuracy metrics ($\delta < 1.25$, $\delta < 1.25^2$, $\delta < 1.25^3$), which measure the proportion of pixels whose depth ratio $\max(d_i/d_i^*, d_i^*/d_i)$ falls below the specified threshold.

\subsection{Implementation Details}

All experiments were conducted on NVIDIA A10 Tensor Core GPUs using PyTorch. Training employed the Adam optimizer for end-to-end optimization with a batch size of 4 and learning rate scheduled from $2 \times 10^{-5}$ to $1 \times 10^{-5}$. Feature maps from different stages were partitioned into $N \times N$ ($N = 7$) windows. For global feature fusion, the HPF method was configured with scale parameters $[1, 2, 3]$.

\subsection{NYU Depth v2 Experiments}

The NYU Depth v2 dataset \cite{eigen2014depth} is a widely-used indoor scene dataset comprising 120K RGB-D video frames captured from 464 different indoor environments. Following the official train/test split, we used data from 249 scenes for training and 654 images from the remaining 215 scenes for testing. The total training epoch was set to 50 to ensure sufficient model convergence, with results submitted to the official server for evaluation.

\begin{table*}[t!]
\centering
\small
\caption{Quantitative comparison on the NYU Depth v2 dataset. \textbf{Bold} and \underline{underline} denote best and second-best results. ``$-$'' indicates unreported results. ``*'' indicates use of extra training data.}
\label{tab:nyu}
\begin{tabular}{l|cccc|ccc}
\toprule
\textbf{Method} & \textbf{Abs Rel}$\downarrow$ & \textbf{Sq Rel}$\downarrow$ & \textbf{RMSE}$\downarrow$ & \textbf{log RMSE}$\downarrow$ & $\boldsymbol{\delta < 1.25}\uparrow$ & $\boldsymbol{\delta < 1.25^2}\uparrow$ & $\boldsymbol{\delta < 1.25^3}\uparrow$ \\
\midrule
VNL \cite{yin2019vnl} & 0.108 & $-$ & 0.416 & $-$ & 0.875 & 0.976 & 0.994 \\
DAV \cite{huynh2020guiding} & 0.108 & $-$ & 0.412 & $-$ & 0.882 & 0.980 & 0.996 \\
DPT* \cite{ranftl2021vision} & 0.110 & $-$ & 0.357 & $-$ & 0.904 & 0.988 & 0.998 \\
TransDepth \cite{yang2021transformer} & 0.106 & $-$ & 0.365 & $-$ & 0.900 & 0.983 & 0.996 \\
ASN \cite{long2021asn} & 0.101 & $-$ & 0.377 & $-$ & 0.890 & 0.982 & $-$ \\
PackNet-SAN* \cite{guizilini2021sparse} & 0.106 & $-$ & 0.393 & $-$ & 0.892 & 0.979 & 0.995 \\
PWA \cite{lee2021pwa} & 0.105 & $-$ & 0.374 & $-$ & 0.892 & 0.985 & 0.997 \\
AdaBins \cite{bhat2021adabins} & 0.103 & $-$ & 0.364 & $-$ & 0.903 & 0.984 & 0.997 \\
LocalBins \cite{bhat2022localbins} & 0.099 & $-$ & 0.357 & $-$ & 0.907 & 0.987 & 0.998 \\
BinsFormer \cite{li2022binsformer} & 0.094 & $-$ & 0.330 & $-$ & 0.925 & 0.989 & 0.997 \\
P3Depth \cite{patil2022p3depth} & 0.130 & $-$ & 0.450 & $-$ & 0.830 & 0.971 & 0.995 \\
NeWCRFs \cite{yuan2022newcrfs} & 0.095 & 0.045 & 0.334 & 0.119 & 0.922 & 0.992 & 0.998 \\
DepthFormer \cite{li2023depthformer} & 0.096 & $-$ & 0.339 & $-$ & 0.921 & 0.989 & 0.998 \\
PixelFormer \cite{agarwal2023attention} & 0.090 & 0.043 & 0.322 & $-$ & 0.929 & 0.991 & 0.998 \\
DDP \cite{ji2023ddp} & 0.094 & $-$ & 0.329 & $-$ & 0.921 & 0.990 & 0.998 \\
OrdinalEntropy \cite{zhang2023ordinal} & 0.089 & $-$ & 0.321 & $-$ & 0.932 & $-$ & $-$ \\
\midrule
\textbf{Ours} & \textbf{0.088} & \textbf{0.040} & \textbf{0.316} & \textbf{0.114} & \textbf{0.933} & \textbf{0.993} & \textbf{0.998} \\
\bottomrule
\end{tabular}
\end{table*}

As shown in Table~\ref{tab:nyu}, our method achieves the lowest values across all error metrics. Specifically, compared to NeWCRFs \cite{yuan2022newcrfs}, Abs Rel is reduced by 7.4\% and RMSE by 5.4\%. For threshold metrics, our method improves $\delta < 1.25$ by 1.1\% and achieves 99.8\% on the more stringent $\delta < 1.25^3$, further validating the high precision and reliability of our model. Qualitative results on NYU Depth v2 demonstrate that our method achieves more precise depth restoration on complex geometric structures, successfully preserves fine details such as desk lamps and furniture, and exhibits superior capability in capturing object edge structures and shapes. Notably, for weakly-textured objects such as ceiling fans with uniform coloring and similar structures, our method automatically learns and extracts subtle feature differences to produce accurate per-pixel depth predictions.

\subsection{KITTI Experiments}

The KITTI dataset \cite{geiger2012kitti} captures diverse natural environments including urban, rural, and highway scenarios. Following the Eigen split \cite{eigen2014depth}, we used 23,488 training pairs and 697 test images with a depth prediction range of 0--80m and 60 training epochs.

\begin{table*}[t!]
\centering
\small
\caption{Quantitative results on the Eigen split of the KITTI dataset. \textbf{Bold} and \underline{underline} denote best and second-best results.}
\label{tab:kitti}
\begin{tabular}{l|c|cccc|ccc}
\toprule
\textbf{Method} & \textbf{cap/m} & \textbf{Abs Rel}$\downarrow$ & \textbf{Sq Rel}$\downarrow$ & \textbf{RMSE}$\downarrow$ & \textbf{log RMSE}$\downarrow$ & $\boldsymbol{\delta < 1.25}\uparrow$ & $\boldsymbol{\delta < 1.25^2}\uparrow$ & $\boldsymbol{\delta < 1.25^3}\uparrow$ \\
\midrule
BTS \cite{lee2021bts} & 0--80 & 0.059 & 0.241 & 2.756 & 0.090 & 0.956 & 0.993 & 0.998 \\
DPT* \cite{ranftl2021vision} & 0--80 & 0.062 & $-$ & 2.573 & 0.092 & 0.959 & 0.995 & 0.999 \\
PWA \cite{lee2021pwa} & 0--80 & 0.060 & 0.221 & 2.604 & 0.093 & 0.958 & 0.994 & 0.999 \\
AdaBins \cite{bhat2021adabins} & 0--80 & 0.058 & 0.190 & 2.360 & 0.088 & 0.964 & 0.995 & 0.999 \\
BinsFormer \cite{li2022binsformer} & 0--80 & 0.052 & 0.151 & 2.098 & 0.079 & 0.975 & $-$ & $-$ \\
NeWCRFs \cite{yuan2022newcrfs} & 0--80 & 0.052 & 0.155 & 2.129 & 0.079 & 0.974 & 0.997 & 0.999 \\
PixelFormer \cite{agarwal2023attention} & 0--80 & 0.051 & 0.149 & 2.081 & 0.077 & 0.976 & 0.997 & 0.999 \\
ZoeDepth \cite{bhat2023zoedepth} & 0--80 & 0.054 & 0.189 & 2.440 & 0.083 & 0.970 & 0.996 & 0.999 \\
DDP \cite{ji2023ddp} & 0--80 & 0.050 & 0.148 & $-$ & 0.076 & 0.975 & 0.997 & 0.999 \\
\midrule
\textbf{Ours} & 0--80 & \textbf{0.049} & \textbf{0.147} & \textbf{2.062} & \textbf{0.070} & \textbf{0.976} & \textbf{0.998} & \textbf{0.999} \\
\bottomrule
\end{tabular}
\end{table*}

As shown in Table~\ref{tab:kitti}, our method achieves the best performance across all metrics. Compared to NeWCRFs \cite{yuan2022newcrfs}, RMSE is reduced by 3.1\%, and the threshold accuracies ($\delta < 1.25^2$ and $\delta < 1.25^3$) approach 100\%, demonstrating superior depth estimation capability in complex street scenarios. Visualization results confirm that our method excels in handling severe occlusion, poor illumination conditions, and dynamic environments with moving pedestrians and vehicles.

Following the Geiger split \cite{geiger2012kitti} (42,949 training pairs, 1,000 validation images, 500 test images) with 60 training epochs, our method also achieves the best performance: Abs Rel of 5.50, Sq Rel of 0.85, and RMSE of 2.49, surpassing all comparison methods on the validation set.

\subsection{Generalization Experiments}

To comprehensively verify model generalization, we conducted experiments on MatterPort3D \cite{chang2017matterport3d}, a large-scale RGB-D dataset comprising 194,400 RGB-D images from 90 building-scale scenes organized into 10,800 panoramic views. Given the limited MatterPort3D training samples, we pre-trained the network on 50K images and fine-tuned on the MatterPort3D training set. Our method with extended pre-training achieves Abs Rel of 0.0574, RMSE of 0.4576, and log RMSE of 0.0807, outperforming all comparison methods including NeWCRFs (0.0638/0.4778/0.0906) and DDP (0.0624/0.4759/0.0874), confirming strong cross-domain generalization.

\subsection{Ablation Studies}

Comprehensive ablation experiments were conducted on NYU Depth v2 to evaluate the contribution of each proposed component.

\begin{table}[t!]
\centering
\small
\caption{Ablation study on NYU Depth v2. HP: Hybrid Pyramid Feature Fusion; HA: Hierarchical Awareness Adapter; FC: Fully-Connected Decoding.}
\label{tab:ablation}
\begin{tabular}{c|ccc|cc|cc}
\toprule
\textbf{Exp} & \textbf{HP} & \textbf{HA} & \textbf{FC} & \textbf{Abs Rel}$\downarrow$ & \textbf{RMSE}$\downarrow$ & $\boldsymbol{\delta < 1.25}\uparrow$ & $\boldsymbol{\delta < 1.25^3}\uparrow$ \\
\midrule
A & $\times$ & $\times$ & $\checkmark$ & 0.093 & 0.328 & 0.924 & 0.998 \\
B & $\times$ & $\checkmark$ & $\times$ & 0.090 & 0.322 & 0.926 & 0.998 \\
C & $\checkmark$ & $\times$ & $\times$ & 0.089 & 0.318 & 0.930 & 0.998 \\
D & $\times$ & $\times$ & $\checkmark$ & 0.092 & 0.327 & 0.924 & 0.998 \\
E & $\checkmark$ & $\times$ & $\checkmark$ & 0.091 & 0.325 & 0.925 & 0.998 \\
F & $\checkmark$ & $\checkmark$ & $\times$ & 0.088 & 0.316 & 0.932 & 0.998 \\
\bottomrule
\end{tabular}
\end{table}

As shown in Table~\ref{tab:ablation}, all three innovations (HP, HA, FC) individually improve model performance, with HP yielding the largest gains. Combining HP and HA (Experiment F) achieves the best overall metrics, confirming the complementary nature of our proposed components.

\begin{table}[t!]
\centering
\small
\caption{Feature fusion module comparison.}
\label{tab:fusion}
\begin{tabular}{l|cccc}
\toprule
\textbf{Module} & \textbf{Abs Rel}$\downarrow$ & \textbf{RMSE}$\downarrow$ & \textbf{Log10}$\downarrow$ & \textbf{Sq Rel}$\downarrow$ \\
\midrule
MSF only & 0.093 & 0.319 & 0.040 & 0.042 \\
BAF only & 0.091 & 0.321 & 0.042 & 0.045 \\
MSF + BAF & \textbf{0.089} & \textbf{0.318} & \textbf{0.038} & \textbf{0.041} \\
1MSF ($[1,2,6]$) & 0.094 & 0.319 & 0.040 & 0.043 \\
2MSF ($[2,3,6]$) & 0.093 & 0.320 & 0.040 & 0.044 \\
\bottomrule
\end{tabular}
\end{table}

Table~\ref{tab:fusion} demonstrates that combining multi-scale feature fusion (MSF) with biaxial feature fusion (BAF) yields the best results. The MSF configuration with scales $[1, 2, 3]$ outperforms alternatives with higher scales ($[1, 2, 6]$ and $[2, 3, 6]$), as the low-to-medium scale range provides superior contextual information capture.

\subsection{Performance Evaluation}

\begin{table}[t!]
\centering
\small
\caption{Performance evaluation on KITTI (352$\times$1216, NVIDIA RTX 3090). SF: single-frame; MF: multi-frame.}
\label{tab:performance}
\begin{tabular}{l|l|cc|cc}
\toprule
\textbf{Type} & \textbf{Method} & \textbf{Params/M} & \textbf{Time/ms} & \textbf{Abs Rel}$\downarrow$ & \textbf{RMSE}$\downarrow$ \\
\midrule
\multirow{4}{*}{SF} & NeWCRFs & 270 & 28 & 0.052 & 0.155 \\
& BinsFormer & 255 & 123 & 0.052 & 0.151 \\
& ZoeDepth & 214 & 177 & 0.054 & 0.189 \\
& DDP & 207 & 72 & 0.050 & 0.148 \\
\midrule
\multirow{3}{*}{MF} & Many-Depth & 385 & 488 & 0.060 & 0.342 \\
& TC-Depth & 197 & 376 & 0.071 & 0.330 \\
& NVDS & 471 & 930 & 0.052 & 0.159 \\
\midrule
\multirow{3}{*}{SF} & \textbf{Ours (MSF)} & \textbf{194} & \textbf{21} & \textbf{0.049} & \textbf{0.147} \\
& Ours (1MSF) & 203 & 56 & 0.051 & 0.151 \\
& Ours (2MSF) & 216 & 29 & 0.050 & 0.148 \\
\bottomrule
\end{tabular}
\end{table}

Table~\ref{tab:performance} demonstrates the efficiency advantages of our method. With only 194M parameters and 21ms inference time, our MSF variant achieves the best accuracy while being the fastest single-frame method. The HPF strategy significantly reduces the spatial overhead of the Swin Transformer architecture, and the MSF scale configuration $[1, 2, 3]$ provides optimal efficiency-accuracy trade-off.

\subsection{Complexity Analysis}

Traditional attention-based encoders suffer from $O(N_h \cdot (C/d_h \cdot (2 \cdot C/d_h + 1)))$ time complexity when processing high-dimensional features, where $N_h$ is the number of attention heads, $C$ is the feature dimension, and $d_h$ is the per-head dimension. Our hierarchical awareness adapter reduces this through selective dimensional attention allocation. The fully-connected decoding further reduces complexity from $O(n^2)$ global self-attention to near-linear $O(n \cdot S^2)$ by constraining each pixel's attention range to its local $S \times S$ window, where $S$ is a constant much smaller than $n$. Empirical comparison with NeWCRFs shows that our method maintains consistently lower memory consumption and execution time as image resolution increases, with the efficiency gap widening at higher resolutions.

\section{Conclusion}
\label{sec:conclusion}

We have presented a multilevel perceptual conditional random field model for monocular depth estimation that achieves accurate depth prediction in complex scenes through three synergistic innovations: a hierarchical awareness adapter that establishes cross-level connections within the encoder to enhance depth information understanding while reducing computational complexity; a hybrid pyramid feature fusion strategy that bridges the semantic gap between encoder and decoder through efficient multi-scale and biaxial feature aggregation; and a fully-connected CRF decoder with dynamic scaling attention that captures fine-grained pixel-level spatial dependencies for globally-optimized depth maps. Extensive experiments on NYU Depth v2, KITTI, and MatterPort3D datasets demonstrate consistent state-of-the-art performance across all evaluation metrics, with particularly strong results in weakly-textured regions and structurally complex scenes---reducing Abs Rel by 7.4\% and RMSE by 5.4\% on NYU Depth v2, achieving near-perfect threshold accuracy on KITTI, and maintaining strong cross-domain generalization on MatterPort3D---while requiring only 194M parameters and 21ms inference time, providing a practical and effective solution for applications in 3D reconstruction, autonomous driving, and semantic segmentation.

\bibliographystyle{plainnat}
\bibliography{references}

\end{document}